# SCIENTIFIC DATA

# Data Descriptor: The HAM10000 dataset, a large collection of multi-source dermatoscopic images of common pigmented skin lesions

Philipp Tschandl[1], Cliff Rosendahl[2] & Harald Kittler[1]

Training of neural networks for automated diagnosis of pigmented skin lesions is hampered by the small size and lack of diversity of available datasets of dermatoscopic images. We tackle this problem by releasing the HAM10000 ("Human Against Machine with 10000 training images") dataset. We collected dermatoscopic images from different populations acquired and stored by different modalities. Given this diversity we had to apply different acquisition and cleaning methods and developed semi-automatic workflows utilizing specifically trained neural networks. The final dataset consists of 10015 dermatoscopic images which are released as a training set for academic machine learning purposes and are publicly available through the ISIC archive. This benchmark dataset can be used for machine learning and for comparisons with human experts. Cases include a representative collection of all important diagnostic categories in the realm of pigmented lesions. More than 50% of lesions have been confirmed by pathology, while the ground truth for the rest of the cases was either follow-up, expert consensus, or confirmation by in-vivo confocal microscopy.

| | |
|---|---|
| **Design Type(s)** | database creation objective ● data integration objective ● image format conversion objective |
| **Measurement Type(s)** | skin lesions |
| **Technology Type(s)** | digital curation |
| **Factor Type(s)** | diagnosis ● Diagnostic Procedure ● age ● biological sex ● animal body part |
| **Sample Characteristic(s)** | Homo sapiens ● skin of body |

[1]ViDIR Group, Department of Dermatology, Medical University of Vienna, Vienna 1090, Austria. [2]Faculty of Medicine, University of Queensland, Herston 4006, Austria. Correspondence and requests for materials should be addressed to P.T (email: philipp.tschandl@meduniwien.ac.at).

## Background & Summary

Dermatoscopy is a widely used diagnostic technique that improves the diagnosis of benign and malignant pigmented skin lesions in comparison to examination with the unaided eye[1]. Dermatoscopic images are also a suitable source to train artificial neural networks to diagnose pigmented skin lesions automatically. In 1994, Binder et al.[2] already used dermatoscopic images successfully to train an artificial neural network to differentiate melanomas, the deadliest type of skin cancer, from melanocytic nevi. Although the results were promising, the study, like most earlier studies, suffered from a small sample size and the lack of dermatoscopic images other than melanoma or nevi. Recent advances in graphics card capabilities and machine learning techniques set new benchmarks with regard to the complexity of neural networks and raised expectations that automated diagnostic systems will soon be available that diagnose all kinds of pigmented skin lesions without the need of human expertise[3].

Training of neural-network based diagnostic algorithms requires a large number of annotated images[4] but the number of high quality dermatoscopic images with reliable diagnoses is limited or restricted to only a few classes of diseases.

In 2013 Mendonça et al. made 200 dermatoscopic images available as the PH2 dataset including 160 nevi and 40 melanomas[5]. Pathology was the ground truth for melanomas but not available for most nevi. Because the set is publicly available (http://www.fc.up.pt/addi/) and includes comprehensive metadata it served as a benchmark dataset for studies of the computer diagnosis of melanoma until now.

Accompanying the book Interactive Atlas of Dermoscopy[6] a CD-ROM is commercially available with digital versions of 1044 dermatoscopic images including 167 images of non-melanocytic lesions, and 20 images of diagnoses not covered in the HAM10000 dataset. Although this is one of the most diverse available datasets in regard to covered diagnoses, its use is probably limited because of its constrained accessibility.

The ISIC archive (https://isic-archive.com/) is a collection of multiple databases and currently includes 13786 dermatoscopic images (as of February 12th 2018). Because of permissive licensing (CC-0), well-structured availability, and large size it is currently the standard source for dermatoscopic image analysis research. It is, however, biased towards melanocytic lesions (12893 of 13786 images are nevi or melanomas). Because this portal is the most comprehensive, technically advanced, and accessible resource for digital dermatoscopy, we will provide our dataset through the ISIC archive.

Because of the limitations of available datasets, past research focused on melanocytic lesions (i.e the differentiation between melanoma and nevus) and disregarded non-melanocytic pigmented lesions although they are common in practice. The mismatch between the small diversity of available training data and the variety of real life data resulted in a moderate performance of automated diagnostic systems in the clinical setting despite excellent performance in experimental settings[3,5,7,8]. Building a classifier for multiple diseases is more challenging than binary classification[9]. Currently, reliable multi-class predictions are only available for clinical images of skin diseases but not for dermatoscopic images[10,11].

To boost the research on automated diagnosis of dermatoscopic images we released the HAM10000 ("Human Against Machine with 10000 training images") dataset. The dataset will be provided to the participants of the ISIC 2018 classification challenge hosted by the annual MICCAI conference in Granada, Spain, but will also be available to research groups who do not participate in the challenge. Because we will also use this dataset to collect and provide information on the performance of human expert diagnosis, it could serve as benchmark set for the comparisons of humans and machines in the future. In order to provide more information to machine-learning research groups who intend to use the HAM10000 training set for research we describe the evolution and the specifics of the dataset (Fig. 1) in detail.

## Methods

The 10015 dermatoscopic images of the HAM10000 training set were collected over a period of 20 years from two different sites, the Department of Dermatology at the Medical University of Vienna, Austria, and the skin cancer practice of Cliff Rosendahl in Queensland, Australia. The Australian site stored images and meta-data in PowerPoint files and Excel databases. The Austrian site started to collect images before the era of digital cameras and stored images and metadata in different formats during different time periods.

### Extraction of images and meta-data from PowerPoint files

Each PowerPoint file contained consecutive clinical and dermatoscopic images of one calendar month of clinical workup, where each slide contained a single image and a text-field with a unique lesion identifier. Because of the large amount of data we applied an automated approach to extract and sort those images. We used the Python package python-pptx to access the PowerPoint files and to obtain the content. We iterated through each slide and automatically extracted and stored the source image, the corresponding identifier, and the year of documentation, which was part of the file name.

### Digitization of diapositives

Before the introduction of digital cameras, dermatoscopic images at the Department of Dermatology in Vienna, Austria were stored as diapositives. We digitized the diapositives with a Nikon Coolscan 5000 ED scanner with a two-fold scan with Digital ICE and stored files as JPEG Images (8-bit color depth) in highest quality (300DPI; $15 \times 10$ cm). We manually cropped the scanned images with the lesion centered to 800x600px at 72DPI, and applied manual histogram corrections to enhance visual contrast and color reproduction (Fig. 2).

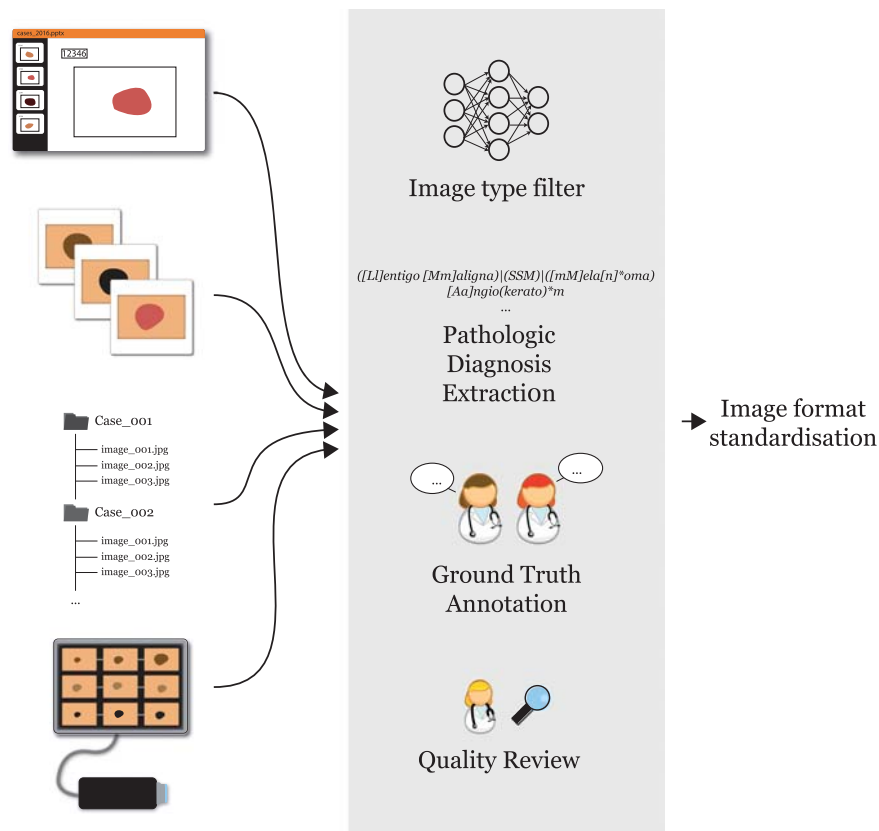

**Figure 1. Schematic flow of dataset workup methods.** Image and data content from different sources were entered into a pipeline to organize and clean data, with final images being standardized and stored in a common format.

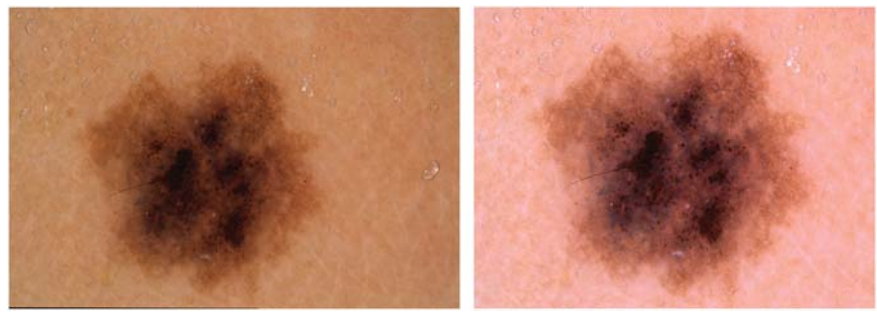

**Figure 2. Manual correction of a scanned diapositive.** Original scanned image (**a**) with remaining black border on the lower left, lesion off center, yellow hue and reduced luminance. On the right (**b**), the final image after manual quality review.

**Extraction of data from a digital dermatoscopy system**

The Department of Dermatology at the University of Vienna is equipped with the digital dermatoscopy system MoleMax HD (Derma Medical Systems, Vienna, Austria). We extracted cases from this system by filtering SQL-tables with a proprietary tool provided by the manufacturer. We selected only non-melanocytic lesions with a consensus benign diagnosis, nevi with >1.5 years of digital dermatoscopic follow-up, and excised lesions with a histopathologic report. Histopathologic reports were matched manually to specific lesions. From a series of multiple sequential images of the same nevus we extracted only the most recent one. Some melanomas of this set were also photographed with a DermLiteTM

FOTO (3GenTM) camera. These additional images became also part of the ViDIR image series, where different images of the same lesion were labeled with a common identifier string. Original images of the MoleMax HD system had a resolution of 1872x1053px (MoleMax HD) with non-quadratic pixels. We manually cropped all MoleMax HD images to 800x600px (72DPI), centered the lesion if necessary, and reverted the format to quadratic pixels.

### Filtering of dermatoscopic images

The source image collections of both sites contained not only dermatoscopic images but also clinical close-ups and overviews. Because there was no reliable annotation of the imaging type, we had to separate the dermatoscopic images from the others. To deal with the large amount of data efficiently we developed an automated method to screen and categorize >30000 images, similar to Han *et al*[12]: We hand-labeled 1501 image files of the Australian collection into the categories "overviews", "close-ups" and "dermatoscopy"[13] (weights pre-trained on ImageNet[4] data) to classify the images according to image type. After training for 20 epochs with Stochastic Gradient Descent, with a learning rate initialized at 0.0003, step-down (Gamma 0.1) at epochs 7 and 13, and a batch-size of 64, we obtained a top-1 accuracy of 98.68% on our hold-out test set. This accuracy was sufficient to accelerate the selection process of dermatoscopic images. The few remaining misclassified close-ups and overviews were removed by hand in a second revision.

### Unifying pathologic diagnoses

Histopathologic diagnoses showed high variability within and between sites including typos, different dermatopathologic terminologies, multiple diagnoses per lesion or uncertain diagnoses. Cases with uncertain diagnoses and collisions were excluded except for melanomas in association with a nevus.

We unified the diagnoses and formed seven generic classes, and specifically avoided ambiguous classifications. The histopathologic expression "superficial spreading melanoma in situ, arising in a preexisting dermal nevus", for example, should only be allocated to the "melanoma" class and not to the nevus class. The seven generic classes were chosen for simplicity and in regard of the intended use as a benchmark dataset for the diagnosis of pigmented lesions by humans and machines. The seven classes covered more than 95% of all pigmented lesions examined in daily clinical practice of the two study sites. A more detailed description of the disease classes is given in the usage notes below.

### Manual quality review

A final manual screening and validation round was performed on all images to exclude cases with the following attributes:

**Type**: Close-up and overview images that were not removed with automatic filtering

**Identifiability**: Images with potentially identifiable content such as garment, jewelry or tattoos

**Quality**: Images that were out of focus or had disturbing artifacts like obstructing gel bubbles. We specifically tolerated the presence of terminal hairs.

**Content**: Completely non-pigmented lesions and ocular, subungual or mucosal lesions

Remaining cases were reviewed for appropriate color reproduction and luminance and, if necessary, corrected via manual histogram correction.

### Code availability

Custom generated code for the described methods is available at https://github.com/ptschandl/HAM10000_dataset.

## Data Records

All data records of the HAM10000 dataset are deposited at the Harvard Dataverse (Data Citation 1). Table 1 shows a summary of the number of images in the HAM10000 training-set according to diagnosis in comparison to existing databases.

Images and metadata are also accessible at the public ISIC-archive through the archive gallery as well as through standardized API-calls (https://isic-archive.com/api/v1).

### Rosendahl image set (Australia)

Lesions of this part of the HAM10000 dataset originate from the office of the skin cancer practice of Cliff Rosendahl (CR, School of Medicine, University of Queensland). We extracted data from his database after institutional ethics board approval (University of Queensland, Protocol-No. 2017001223). Images were solely taken by author CR with either a DermLite Fluid (non-polarized) or DermLite DL3 (3Gen, San Juan Capistrano, CA, USA) with immersion fluid (either 70% ethanol hand-wash gel, or ultrasound gel). Digital images were stored within PowerPoint[TM] presentations. Each slide contained a dermatoscopic image and a text-field with a consecutive unique lesion ID linking the image to a separate Excel database with clinical metadata and histopathologic diagnoses. Images were documented in this manner consecutively starting 2008 until May 2017. The Rosendahl series consists of 34.2GB of digital slides (122 PowerPoint[TM]-files) from which we extracted 36802 images with a matching histopathologic report as described above. After removal of non-pigmented lesions, overviews, close-ups without dermatoscopy, and cases without or with an inappropriate diagnosis that did not fall into one of the predefined generic classes we arrived at the final dataset described in Table 1.

| Dataset | License | Total images | Pathologic verification (%) | akiec | bcc | bkl | df | mel | nv | vasc |
|---------|---------|-------------|----------------------------|-------|-----|-----|-----|-----|-----|------|
| PH2 | Research&Education[a] | 200 | 20.5% | - | - | - | - | 40 | 160 | - |
| Atlas | No license | 1024 | unknown | 5 | 42 | 70 | 20 | 275 | 582 | 30 |
| ISIC 2017[b] | CC-0 | 13786 | 26.3% | 2 | 33 | 575 | 7 | 1019 | 11861 | 15 |
| Rosendahl | CC BY-NC 4.0 | 2259 | 100% | 295 | 296 | 490 | 30 | 342 | 803 | 3 |
| ViDIR Legacy | CC BY-NC 4.0 | 439 | 100% | 0 | 5 | 10 | 4 | 67 | 350 | 3 |
| ViDIR Current | CC BY-NC 4.0 | 3363 | 77.1% | 32 | 211 | 475 | 51 | 680 | 1832 | 82 |
| ViDIR MoleMax | CC BY-NC 4.0 | 3954 | 1.2% | 0 | 2 | 124 | 30 | 24 | 3720 | 54 |
| **HAM10000** | CC BY-NC 4.0 | **10015** | 53.3% | 327 | 514 | 1099 | 115 | 1113 | 6705 | 142 |

**Table 1. Summary of publicly available dermatoscopic image datasets in comparison to HAM10000.**
[a]Not stated specifically on the providing website, we infer their given terms-of-use largely reflect CC BY-NC-ND 3.0. [b]Eight different datasets with CC-0 licensing combined as available on February 12th 2018.

### ViDIR image set (Austria)
From the ViDIR Group (Department of Dermatology at the Medical University of Vienna, Austria) datasources from different times were available and processed after ethics committee approval at the Medical University of Vienna (Protocol-No. 1804/2017).

**Legacy Diapositives.** The oldest resource of images of the ViDIR group dates back to the era before the availablility of digital cameras when dermatoscopic images were taken with analog cameras and archived as diapositives. They were originally photographed with the Heine Dermaphot system using immersion fluid, and produced for educational and archival purposes with the E-6 method[14].

**Current Database.** Since 2005 we documented digital dermatoscopic images with a DermLiteTM FOTO (3GenTM) system (single cases also with Heine Delta 20) and stored images and meta-data in a central database. This dataset includes triplets from the same lesion taken with different magnifications to enable visualization of local features and general patterns.

**MoleMax Series.** The Department of Dermatology in Vienna offers digital dermatoscopic follow-up to high risk patients to increase the the number of excisions of early melanoma while decreasing unnecessary excisions of nevi[15]. The time interval between follow-up visits usually ranges from 6-12 months. Rinner et al.[16] recently published a detailed description of this follow-up program. Since 2015 the MoleMax HD System (Derma Medical Systems, Vienna, Austria) is used for acquisition and storage of dermatoscopic images.Most patients in the follow-up program have multiple nevi. Dermatologists of the Department of Dermatology in Vienna usually monitor "atypical" nevi but also a small number of randomly selected inconspicuous nevi, which are commonly underrepresented in datasets used for machine-learning.

### Technical Validation
If necessary, an expert dermatologist performed manual histogram correction to adjust images for visual inspection by human readers. We restricted corrections to underexposed images and images with a visible yellow or green hue and made no other adjustments. To illustrate changes, we conducted color illuminant estimations after corrections according to the grey-world assumption[17] as shown in Fig. 3. As desired, corrections shifted yellow and green illuminants towards blue and red. We defined four different types of ground truth:

#### Histopathology
Histopathologic diagnoses of excised lesions have been performed by specialized dermatopathologists. We scanned all available histopathologic slides of the current ViDIR image set for later review. We manually reviewed all images with the corresponding histopathologic diagnosis and checked for plausibility. If the histopathologic diagnosis was implausible we checked for sample mismatch and reviewed the report and reexamined the slide if necessary. We excluded cases with ambiguous histopathologic diagnoses (for example: "favor nevus but cannot rule out evolving melanoma in situ").

#### Confocal
Reflectance confocal microscopy is an in-vivo imaging technique with a resolution at near-cellular level[18], and some facial benign keratoses were verified by this method. Most cases were included in a prospective confocal study conducted at the Department of Dermatology at the Medical University of Vienna that also included follow-up for one year[19].

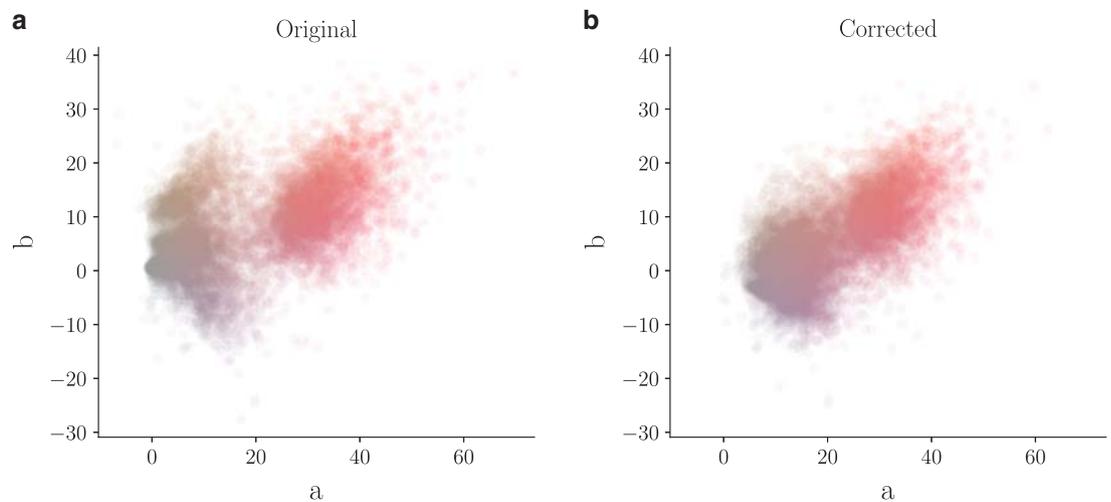

**Figure 3. Illuminant color estimation of the dataset.** Values estimated with a grey-world assumption of all training-set images in Lab-color space before (left) and after (right) manual histogram changes.

**Follow-up**

If nevi monitored by digital dermatoscopy did not show any changes during 3 follow-up visits or 1.5 years we accepted this as evidence of biologic benignity. Only nevi, but no other benign diagnoses were labeled with this type of ground-truth because dermatologists usually do not monitor dermatofibromas, seborrheic keratoses, or vascular lesions. The presence of change was assessed by author HK who has more than 20 years of experience in digital dermatoscopic follow-up.

**Consensus**. For typical benign cases without histopathology or follow-up we provide an expert-consensus rating of authors PT and HK. We applied the consensus label only if both authors independently gave the same unequivocal benign diagnosis. Lesions with this type of ground-truth were usually photographed for educational reasons and did not need further follow-up or biopsy for confirmation.

## Usage Notes

The HAM10000 training set includes pigmented lesions from different populations. The Austrian image set consists of lesions of patients referred to a tertiary European referral center specialized for early detection of melanoma in high risk groups. This group of patients often have a high number of nevi and a personal or family history of melanoma. The Australian image set includes lesions from patients of a primary care facility in a high skin cancer incidence area. Australian patients are typified by severe chronic sun damage. Chronic sun damaged skin is characterized by multiple solar lentigines and ectatic vessels, which are often present in the periphery of the target lesion. Very rarely also small angiomas and seborrheic keratoses may collide with the target lesion. We did not remove this "noise" and we also did not remove terminal hairs because it reflects the situation in clinical practice. In most cases, albeit not always, the target lesion is in the center of the image. Dermatoscopic images of both study sites were taken by different devices using polarized and non-polarized dermatoscopy. The set includes representative examples of pigmented skin lesions that are practically relevant. More than 95% of all lesion encountered during clinical practice will fall into one of the seven diagnostic categories. In practice, the task of the clinician is to differentiate between malignant and benign lesions, but also to make specific diagnoses because different malignant lesions, for example melanoma and basal cell carcinoma, may be treated in a different way and timeframe. With the exception of vascular lesions, which are pigmented by hemoglobin and not by melanin, all lesions have variants that are completely devoid of pigment (for example amelanotic melanoma). Non-pigmented lesions, which are more diverse and have a larger number of possible differential diagnoses, are not part of this set.

The following description of diagnostic categories is meant for computer scientists who are not familiar with the dermatology literature:

**akiec**

Actinic Keratoses (Solar Keratoses) and Intraepithelial Carcinoma (Bowen's disease) are common non-invasive, variants of squamous cell carcinoma that can be treated locally without surgery. Some authors regard them as precursors of squamous cell carcinomas and not as actual carcinomas. There is, however, agreement that these lesions may progress to invasive squamous cell carcinoma – which is usually not pigmented. Both neoplasms commonly show surface scaling and commonly are devoid of pigment.

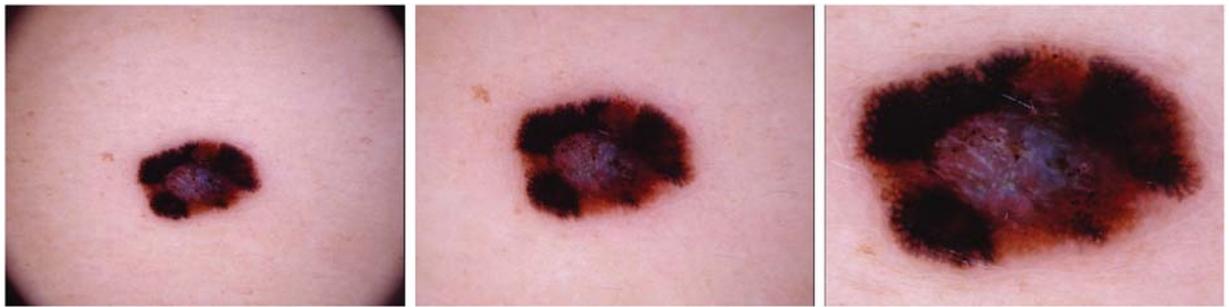

**Figure 4. Example triplet of the same lesion.** Rather than cropped to different sizes of a large source image, these were originally taken at different magnifications and angles.

Actinic keratoses are more common on the face and Bowen's disease is more common on other body sites. Because both types are induced by UV-light the surrounding skin is usually typified by severe sun damaged except in cases of Bowen's disease that are caused by human papilloma virus infection and not by UV. Pigmented variants exist for Bowen's disease[20] and for actinic keratoses[21], and both are included in this set.

The dermatoscopic criteria of pigmented actinic keratoses and Bowen's disease are described in detail by Zalaudek et al.[22,23] and by Cameron et al.[20].

**bcc**
Basal cell carcinoma is a common variant of epithelial skin cancer that rarely metastasizes but grows destructively if untreated. It appears in different morphologic variants (flat, nodular, pigmented, cystic), which are described in more detail by Lallas et al.[24].

**bkl**
"Benign keratosis" is a generic class that includes seborrheic keratoses ("senile wart"), solar lentigo - which can be regarded a flat variant of seborrheic keratosis - and lichen-planus like keratoses (LPLK), which corresponds to a seborrheic keratosis or a solar lentigo with inflammation and regression[25]. The three subgroups may look different dermatoscopically, but we grouped them together because they are similar biologically and often reported under the same generic term histopathologically. From a dermatoscopic view, lichen planus-like keratoses are especially challenging because they can show morphologic features mimicking melanoma[26] and are often biopsied or excised for diagnostic reasons. The dermatoscopic appearance of seborrheic keratoses varies according to anatomic site and type[27].

**df**
Dermatofibroma is a benign skin lesion regarded as either a benign proliferation or an inflammatory reaction to minimal trauma. The most common dermatoscopic presentation is reticular lines at the periphery with a central white patch denoting fibrosis[28].

**nv**
Melanocytic nevi are benign neoplasms of melanocytes and appear in a myriad of variants, which all are included in our series. The variants may differ significantly from a dermatoscopic point of view. In contrast to melanoma they are usually symmetric with regard to the distribution of color and structure[29].

**mel**
Melanoma is a malignant neoplasm derived from melanocytes that may appear in different variants. If excised in an early stage it can be cured by simple surgical excision. Melanomas can be invasive or non-invasive (in situ). We included all variants of melanoma including melanoma in situ, but did exclude non-pigmented, subungual, ocular or mucosal melanoma.

Melanomas are usually, albeit not always, chaotic, and some melanoma specific criteria depend on anatomic site[23,30].

**vasc**
Vascular skin lesions in the dataset range from cherry angiomas to angiokeratomas[31] and pyogenic granulomas[32]. Hemorrhage is also included in this category.

Angiomas are dermatoscopically characterized by red or purple color and solid, well circumscribed structures known as red clods or lacunes.

The number of images in the datasets does not correspond to the number of unique lesions, because we also provide images of the same lesion taken at different magnifications or angles (Fig. 4), or with different cameras. This should serve as a natural data-augmentation as it shows random transformations and visualizes both general and local features.

Images in the dataset may diverge from those an end-user, especially lay persons, would provide in a real-world scenario. For example, insufficiently magnified images of small lesions or out of focus images were removed.

We used automated screening by neural networks, multiple manual reviews and cleared EXIF-data of the images to remove any potentially identifiable information. Data can thus be regarded anonymized to the best of our knowledge and the data collection was approved by the ethics review committee at the Medical University of Vienna and the University of Queensland.

## Data Citation

## Acknowledgements

We want to thank Andreas Ebner (photographer at the Department of Dermatology, Medical University of Vienna) for archival and acquisition of most ViDIR dataset images. We would further thank Giuliana

Petronio for scanning the diapositives and Christoph Sinz, M.D. for invaluable help on continuously curating the ViDIR databases. We are grateful to our collaborators of the ISIC-Archive and ISIC 2018 challenge: M. Emre Celebi Ph.D. (University of Central Arkansas, Arkansas, USA), Noel C. F. Codella Ph. D. (IBM Research, New York, USA), Kristin Dana Ph.D. (Rutgers University, New Jersey, USA), David Gutman M.D. (Emory University, Georgia, USA), Allan Halpern M.D. (Memorial Sloan Kettering Cancer Center, New York, USA) and Brian Helba (Kitware, New York, USA). They made hosting and immediate application of the dataset possible and provided valuable input during workup of the data.

## Author Contributions

P.T. wrote the Data Descriptor, and performed data handling, image processing and analysis, expert annotation and quality review. C.R. wrote the Data Descriptor, and collected all cases from the Rosendahl-Series. H.K. wrote the Data Descriptor, and performed image annotation, extraction of MoleMax Series, and quality review of all images.

## Additional Information

**Competing interests**: The authors declare no competing interests.

**How to cite this article**: Tschandl, P. *et al.* The HAM10000 dataset, a large collection of multi-source dermatoscopic images of common pigmented skin lesions. *Sci. Data* 5:180161 doi: 10.1038/sdata.2018.161 (2018).

**Publisher's note**: Springer Nature remains neutral with regard to jurisdictional claims in published maps and institutional affiliations.